\documentclass[runningheads]{llncs}
\usepackage{graphicx}
\usepackage{mathrsfs}
\usepackage{dsfont}
\usepackage{amssymb}
\usepackage{multirow}

\newcommand{\tabincell}[2]{\begin{tabular}{@{}#1@{}}#2\end{tabular}}

\begin{document}
%
\title{Self-supervised Feature Learning for 3D Medical Images by Playing a Rubik's Cube}
%
%
\author{Xinrui Zhuang\inst{1, 2}\thanks{This work was done when Xinrui Zhuang was an intern at YouTu Lab} \and
Yuexiang Li\inst{2} \and
Yifan Hu\inst{2} \and Kai Ma\inst{2} \and Yujiu Yang\inst{1} \and Yefeng Zheng\inst{2}
}

\authorrunning{X. Zhuang et al.}
%
\institute{Graduate School at Shenzhen, Tsinghua University, Shenzhen, China
\email{yang.yujiu@sz.tsinghua.edu.cn} \and
YouTu Lab, Tencent, Shenzhen, China\\
\email{vicyxli@tencent.com}
}
\maketitle              
\begin{abstract}
  Witnessed the development of deep learning, increasing number of studies try to build computer aided diagnosis systems for 3D volumetric medical data. However, as the annotations of 3D medical data are difficult to acquire, the number of annotated 3D medical images is often not enough to well train the deep learning networks. The self-supervised learning deeply exploiting the information of raw data is one of the potential solutions to loose the requirement of training data. In this paper, we propose a self-supervised learning framework for the volumetric medical images. A novel proxy task, i.e., Rubik's cube recovery, is formulated to pre-train 3D neural networks. The proxy task involves two operations, i.e., cube rearrangement and cube rotation, which enforce networks to learn translational and rotational invariant features from raw 3D data. Compared to the train-from-scratch strategy, fine-tuning from the pre-trained network leads to a better accuracy on various tasks, e.g., brain hemorrhage classification and brain tumor segmentation. We show that our self-supervised learning approach can substantially boost the accuracies of 3D deep learning networks on the volumetric medical datasets without using extra data. To our best knowledge, this is the first work focusing on the self-supervised learning of 3D neural networks.

\end{abstract}
\section{Introduction}
Compared with natural images, most medical images, e.g. computed tomography (CT) and magnetic resonance imaging (MRI), are volumetric which appear in a 3D form. A traditional diagnosis approach requires experienced physicians to manually browse the 3D volume data and search for the traits of abnormality, which is laborious and suffers from the problem of inter-observer variation. Due to the development of deep learning, researchers proposed various 3D network architectures \cite{cicek_3d_2016} to assist physicians in increasing the diagnosis accuracy. However, the training of deep learning models may require a large amount of training data. As the annotations of 3D medical images are difficult to acquire, i.e., each 3D volume requires experienced physicians to spend a couple of hours or even days for investigation, the performance of 3D deep learning frameworks suffers from the limited amount of annotated medical images.

To deal with the deficient annotated data, researchers attempted to exploit useful information from the unlabeled data with unsupervised approaches \cite{Spitzer_2018,Zhang_2017}. More recently, the self-supervised learning, as a new paradigm of unsupervised learning, attracts increasing attentions from the community. The pipeline consists of two steps: 1) pre-train a convolutional neural network (CNN) on a proxy task with a large non-annotated dataset. 2) fine-tune the pre-trained network for the specific target task with a small set of annotated data. The proxy task enforces neural networks to deeply mine useful information from the unlabeled raw data, which can boost the accuracy of the subsequent target task with limited training data. Various proxy tasks had been proposed, which include grayscale image colorization \cite{larsson_colorization_2017},  jigsaw puzzle \cite{NorooziVFP18}, object motion estimation \cite{lee_unsupervised_2017} and rotation prediction \cite{Gidaris18}.

For the applications with medical data, researchers took some prior-knowledge into account when formulating the proxy task. Zhang et al. \cite{Zhang_2017} defined a proxy task that sorted the 2D slices extracted from the conventional 3D CT and MR volumes, to pre-train the neural networks for the fine-grained body part recognition (the target task). Spitzer et al. \cite{Spitzer_2018} proposed to pre-train neural networks on a self-supervised learning task, i.e., predicting the 3D distance between two patches sampled from the same brain, for the better segmentation of brain areas (the target task). However, all of the aforementioned self-supervised learning frameworks \cite{Spitzer_2018,Zhang_2017}, including those for natural images \cite{larsson_colorization_2017,NorooziVFP18,lee_unsupervised_2017}, were proposed for 2D networks. As the 3D neural networks integrating the 3D spatial information usually outperform the 2D networks on volumetric medical data, a 3D-based self-supervised learning approach is worthwhile to develop.

In this paper, we propose a 3D-based self-supervised learning approach for volumetric medical data. We formulate a novel proxy task, namely Rubik's cube recovery, to deeply exploit the rich information from 3D medical data and loose the requirement of training data to well train a 3D deep learning model. Like playing a Rubik's cube, there are two operations in the process of our Rubik's cube recovery, i.e., cube rearrangement and cube rotation, which enforce the network to learn the features invariant to translation and rotation from the raw data. The pre-trained 3D network is then fine-tuned on two target tasks, i.e., brain hemorrhage classification and brain tumor segmentation. Experimental results show that the proposed approach can significantly improve accuracy of the 3D CNNs on target tasks, although the model is never explicitly pre-trained to exploit knowledge of brain hemorrhage and tumors. To our best knowledge, this is the first work focusing on the self-supervised learning of 3D CNNs.

\section{Method}
\begin{figure}[!tb]
  \includegraphics[width=\textwidth]{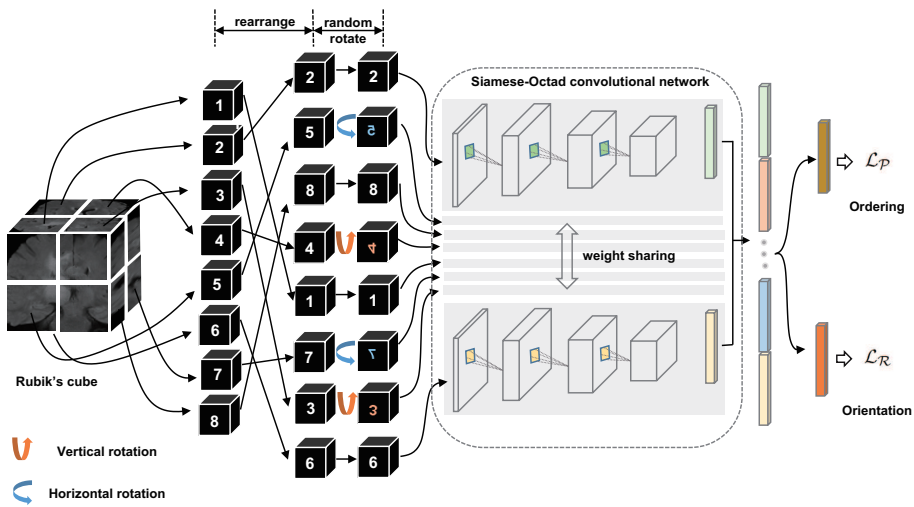}
  \caption{Rubik's cube recovery. The proxy task has two operations, i.e., cube rearrangement and cube rotation.} \label{fig1}
\end{figure}

In this section, we introduce the proposed 3D self-supervised learning approach in details. The proposed approach aims to address the problem of deficient annotated 3D medical data by deeply exploiting the useful information from the limited training data. The approach first pre-trains a 3D CNN on the proxy task and then fine-tunes the pre-trained weights on the target tasks with manual annotations. Inspired by the jigsaw puzzle \cite{NorooziVFP18}, a novel proxy task (Rubik's cube recovery), is proposed for the 3D neural networks. The pipeline of the proxy task is illustrated in Fig.~\ref{fig1}.

\subsection{Rubik's Cube Recovery}
For a 3D medical volume, we first partition it into a grid (e.g., $2 \times 2 \times 2$) of cubes, and then permute the cubes with random rotations. Like playing a Rubik's cube, the proxy task aims to recover the original configuration, i.e., cubes are ordered and orientated.

Compared to the jigsaw puzzle, the Rubik's cube recovery task has two main differences: 1) The Rubik's cube recovery works on 3D volumetric data, while the jigsaw puzzle is proposed for 2D natural images; 2) The difficulty of recovering Rubik's cube is increased by adding the cube rotation operation, which encourages deep learning networks to leverage more spatial information.

\subsubsection{Pre-processing.}
The neural networks are encouraged to learn and use high-level semantic features for Rubik's cube recovery rather than the texture information close to the cube boundaries. Therefore, we leave a gap (about 10 voxels) between two adjacent cubes during volume participation. The cube intensities are normalized to [-1, 1] by using the mean and maximum intensity.

\subsubsection{Network architecture.}
As Fig.~\ref{fig1} shows, a Siamese network with $M$ (which is the number of cubes) sharing weight branches, namely Siamese-Octad, is adopted to solve Rubik's cube. The backbone network for each branch can be any widely-used 3D CNN, e.g., 3D VGG \cite{Simonyan15}. The feature maps from the last fully-connected or convolution layer of all branches are concatenated and given as input to the fully-connected layer of separate tasks, i.e., cube ordering and orientating, which are supervised by permutation loss ($\mathcal{L}_{P}$) and rotation loss ($\mathcal{L}_{R}$), respectively.

\subsubsection{Cube ordering.}
The first step of our Rubik's cube recovery is the cube rearrangement. Taking a 2nd-order Rubik's cube, i.e., $2 \times 2 \times 2$ shown in Fig.~\ref{fig1}, as an example, we first yield all the permutations ($\mathds{P}$) of cubes, i.e., $\mathds{P}= (P_{1}, P_{2},...,P_{8!})$. The permutations control the ambiguity of the task, if two permutations are too close to each other, the Rubik's cube recovery task becomes challenging and ambiguous for networks to learn. Therefore, we iteratively select the $ K $ permutations with the largest Hamming distance from $\mathds{P}$. Then, for each time of Rubik's cube recovery, the eight cubes are rearranged according to one of the $ K $ permutations, e.g., $(2, 5, 8, 4, 1, 7, 3, 6) $ in Fig.~\ref{fig1}. To properly reorder the cubes, the network is trained to identify the selected permutation from the $ K $ options, which can be seen as a classification task with $ K $ categories. Assuming the $1 \times K$ network prediction as $p$ and the one-hot label as $l$, the permutation loss ($\mathcal{L}_{P}$) in this step can be defined as:

\begin{equation}
  \mathcal{L_{P}} = - \sum_{j=1}^{K}{l_{j}\log{p_{j}}} \ .
\end{equation}

\subsubsection{Cube orientation.}
The jigsaw puzzles only involve the translational motion of image tiles on a 2D plane, which makes the network only extract translational invariant features. In our 3D Rubik's cube task, we perform a new operation, i.e., random cube rotation, to encourage network to learn the rotational invariant features as well.

As the cubes often have a cuboid shape, free rotations result in $3 \ (axes) \times 2 \ (directions) \times 4 \ (angles) = 24$ configurations. To reduce the complexity of the task, we limit the directions for cube rotation, i.e., only allowing $180^{\circ}$ horizontal and vertical rotations. As Fig.~\ref{fig1} shows, the cubes (5, 7) and (4, 3) are horizontally and vertically rotated, respectively. To orientate the cubes, the network is required to recognize whether each of the input cubes has been rotated. It can be seen as a multi-label classification task using the $1 \times M$ ($M$ is the number of cubes) ground truth ($g$) with 1 on the positions of rotated cubes and 0 vice versa. Hence, the predictions of this task are two $1 \times M$ vectors ($r$) indicating the possiblities of horizontal ($hor$) and vertical ($ver$) rotations for each cube. The rotation loss ($\mathcal{L}_{R}$) can be written as:

\begin{equation}
  \mathcal{L_{R}} = - \sum_{i=1}^{M}({g_{i}^{hor}\log{r_{i}^{hor}}} + {g_{i}^{ver}\log{r_{i}^{ver}})} \ .
\end{equation}

\subsubsection{Objective.}
With the previously defined permutation loss ($\mathcal{L}_{P}$) and rotation loss ($\mathcal{L}_{R}$), the full objective ($\mathcal{L}$) for our 3D self-supervised CNN is summarized as:

\begin{equation}
  \mathcal{L} = \alpha \mathcal{L}_{P} + \beta \mathcal{L}_{R}  \ .
\end{equation}
where $\alpha$ and $\beta$ are loss weights, ajusting the relative influence of two tasks. We empirically find that equal weights $\alpha = \beta = 0.5$ leads to the best feature representations of pre-trained networks in the experiments.

\subsection{Adapting Pre-trained Weights for Pixel-wise Target Task}
The CNN pre-trained on Rubik's cube recovery task can achieve a robust feature representation, which can then be transferred to the target tasks. For the classification task, the pre-trained CNN can be directly used for finetuning. For the segmentation of 3D medical images, the pre-trained weights can only be adapted to the encoder part of the fully convolutional network (FCN), e.g. U-Net \cite{cicek_3d_2016}. The decoder of FCN still needs random initialization, which may wreck the pre-trained feature representation and neutralize the improvement generated by the pre-training. Inspired by the dense upsampling convolution (DUC) \cite{Wang01}, we propose to apply convolutional operations directly on feature maps yield by the pre-trained encoder to get the dense pixel-wise prediction instead of the transposed convolutions. The DUC can significantly decrease the number of trainable parameters of the decoder and alleviate the influence caused by random initialization.

\section{Experiment}
In this section, we transfer the weights pre-trained on Rubik's cube recovery to two 3D medical image analysis tasks, i.e., pathological cause of brain hemorrhage classification and brain tumor segmentation. The datasets adopted in this study are randomly separated to training and test sets according to the ratio of 80:20.

\subsection{Datasets}
\subsubsection{Brain hemorrhage dataset.} We collected 1486 brain CT scan images from a collaborative hospital, which are used to analyze the pathological cause of brain hemorrhage. The 3D CT volumes containing brain hemorrhage can be classified to four pathological causes, i.e., aneurysm, arteriovenous malformation, moyamoya disease and hypertension. Each 3D CT volume is of size $230 \times 270 \times 30$ voxels. The weight pre-trained on Rubik's cube recovery can be directly transferred to this target task, i.e., brain hemorrhage classification. The cube size of Rubik's cube is $64 \times 64 \times 12$. The average classification accuracy (ACC) is adopted as metric for the performance evaluation.

\subsubsection{BraTS-2018.} The BraTS-2018 training set \cite{Menze_2015} consists of 285 brain tumor MR volumes, which have four modalities, i.e., native T1-weighted (T1), post-contrast T1-weighted (T1Gd), T2-weighted (T2), and T2 Fluid Attenuated Inversion Recovery (FLAIR). All MR images are co-registered to the same anatomical template, interpolated to the same resolution ($1 \ mm^3$) and skull-stripped. The size of each volume is $240 \times 240 \times 155$ voxels. This dataset is widely-used to evaluate the accuracy of segmentation methods for brain tumors. The cube size of Rubik's cube is $64 \times 64 \times 64$. As the BraTS-2018 has four modalities, we concatenate the cubes from different modalities and send to each branch of Siamese-Octad network as input. The mean intersection over union (mIoU) \cite{Garcia_Garcia_2017} is adopted as the metric to evaluate the segmentation accuracy.

\subsection{Performance on Solving Rubik's Cube}
We evaluate the performance of the Siamese-Octad network on Rubik's cube recovery to verify whether the network can deal with the proxy task. The backbone of our Rubik's cube network (Siamese-Octad) is the 3D VGG \cite{Simonyan15}, which is widely-used in self-supervised studies \cite{NorooziVFP18} and 3D medical image processing \cite{cicek_3d_2016}. The test accuracies of $2 \times 2 \times 2$ Rubik's cube recovery on two datasets are listed in Table~\ref{table1}. As the random cube rotation increases the difficulty of solving Rubik's cube, the test accuracies of cube ordering degrade with $-7.7\%$ and $-6.6\%$ for brain hemorrhage dataset and BraTS-2018, respectively. On the other hand, the Rubik's cube network can achieve test accuracies of 93.1\% and 82.1\% for the cube orientation. The experimental results demonstrate that the cube rotation enables networks to develop the concept of rotated content, which means more structural information of brains is extracted compared to the rearrangment-only approach.

\begin{table}[!htb]
  \centering
  \caption{The test accuracies of solving $2 \times 2 \times 2$ Rubik's cube on two datasets.}
  \label{table1}
  \begin{tabular}{c|cc|cc}
    \hline
    \multirow{2}{*}{\bf Dataset} & \multirow{2}{*}{\bf Rearrange} & \multirow{2}{*}{\bf Rotate} & \multicolumn{2}{c}{\bf Accuracy (\%)}               \\ \cline{4-5}
                                 &                                &                             & Ordering                              & Orientation \\ \hline\hline
    Brain hemorrhage dataset     & \checkmark                     & \                           & 99.7                                  & -           \\
    Brain hemorrhage dataset     & \checkmark                     & \checkmark                  & 92.0                                  & 93.1        \\
    BraTS-2018                   & \checkmark                     & \                           & 99.5                                  & -           \\
    BraTS-2018                   & \checkmark                     & \checkmark                  & 92.9                                  & 82.1        \\ \hline
  \end{tabular}
\end{table}

\subsection{Fine-tuning Models on Target Tasks}
We fine-tuned the networks pre-trained on the Rubik's cube recovery for the target tasks to evaluate the benefit produced by pre-trained weights. The training strategies, including train-from-scratch, fine-tuning with weights pre-trained on natural dataset (UCF101 \cite{Khurram_2012}), are involved in comparison experiments. The test results are listed in Table~\ref{table2}.

\subsubsection{Baselines.}
The train-from-scratch strategy is involved as the baseline. Furthermore, similar to the ImageNet pre-trained weights widely-used for 2D image processing, the action recognition dataset, i.e., UCF101, is adopted to pre-train our 3D CNNs. The UCF101 consists of 13320 videos, which can be classified to 101 action categories. We extract frames from videos to form a cube of $112 \times 112 \times 16$ to pre-train the 3D network. The pre-trained models are then transferred to the two target tasks for performance comparison. It is worthwhile to mention that our Rubik's cube pre-trained weights are generated by deeply exploiting useful information from limited training data without using any extra dataset.

\begin{table}[!tb]
  \centering
  \caption{Test accuracies of models with different training strategies on target tasks.}
  \label{table2}
  \begin{tabular}{c|c|c|c}
    \hline
                                       & \tabincell{c}{\bf Brain hemorrhage                                                                 \\\bf cla. (ACC \%)}  & \multicolumn{2}{c}{\tabincell{c}{\bf Brain tumor\\\bf seg. (mIoU \%)}}                                      \\ \cline{2-4}
                                       & {\bf 3D VGG} \cite{Simonyan15}     & {\bf U-Net} \cite{cicek_3d_2016} & {\bf 3D DUC} \cite{Wang01} \\ \hline\hline
    {\bf Train-from-scratch}           & 72.6                               & 73.3                             & 74.0                       \\ \hline
    {\bf Fine-tuned on UCF101}         & 75.3                               & 75.2                             & 76.8                       \\ \hline
    {\bf Cube Ordering}                & 81.1                               & 73.9                             & 75.0                       \\ \hline
    {\bf Rubik's Cube Recovery (Ours)} & {\bf 83.8}                         & {\bf 76.2}                       & {\bf 77.3}                 \\ \hline
  \end{tabular}
\end{table}

\subsubsection{Brain hemorrhage classification.}
As Table~\ref{table2} shows, finetuning from the pre-trained weights can improve the accuracies of models for brain hemorrhage classification, compared to the train-from-scratch. Due to the gap between natural video and volumetric medical data, the improvement yielded by UCF101 pre-trained weights is limited, i.e., $+2.7\%$. In comparison, our Rubik's cube pre-trained weights substantially boost the classification accuracy to 83.8\%, which is 11.2\% higher than that of train-from-scratch model.

\subsubsection{Brain tumor segmentation.}
The mIoU of brain tumors yielded by models trained with different training startegies is also listed in Table~\ref{table2}. Two kinds of FCNs, i.e., U-Net \cite{cicek_3d_2016} and DUC \cite{Wang01}, are involved to evaluate the influence caused by random initialization of decoder. Compared to the models transferred from UCF101 pre-trained weights, the ones fine-tuned from our Rubik's cube recovery paradigm can generate more accurate segmentations for brain tumors, i.e., mIoUs of 76.2\% and 77.3\% are achieved by the U-Net and 3D DUC, respectively.

As the Rubik's cube recovery task only pre-trains the downsampling layers, the decoder (upsampling layers) of U-Net needs to be randomly initialized, which may wreck the feature representations learned by the pre-trained weights and consequently degrade the performance improvement. To alleviate the influence caused by random initialization, the DUC module, which significantly reduces the number of trainable parameters contained in the decoder, is more suitable for the transfer learning on pixel-wise prediction task. It can be observed from Table~\ref{table2} that the 3D DUCs outperform the 3D U-Nets under all pre-training protocols, i.e., $+1.6\%$ and $+1.1\%$ for UCF101 and Rubik's cube pre-trained weights, respectively.

\subsubsection{Comparison of solving different Rubik's cubes.}
Table~\ref{table2} shows the results of models fine-tuned from Rubik's cube without cube rotation as well. The models transferred from our Rubik's cube significantly outperform the ones only pre-trained with cube ordering task, i.e., $+2.7\%$ and $+2.3\%$ for brain hemorrhage classification and brain tumor segmentation, respectively. The experimental result reveals that the difficult Rubik's cube task may lead to the better generalization of models. Although the accuracy of cube ordering decreases by adding the cube rotation (as shown in Table~\ref{table1}), the 3D neural networks pre-trained on the multi-tasks, i.e., cube ordering and orientation, seem to exploit a more robustness feature representation, i.e., translational and rotational invariant, from the raw 3D data.

\section{Conclusion}
In this paper, we proposed a self-supervised learning framework for the volumetric medical images. A novel proxy task, i.e., Rubik's cube recovery, was formulated to pre-train 3D neural networks. The proxy task involved two operations, i.e., cube rearrangement and cube rotation, which enforced networks to learn translational and rotational invariant features from raw 3D data.

\section*{Acknowledge}
The work was supported by the National Key Research and Development Program of China (No. 2018YFB1601102), the Natural Science Foundation of China (No. 61702339), the Key Area Research and Development Program of Guangdong Province, China (No. 2018B010111001), and Shenzhen special fund for the strategic development of emerging industries (No. JCYJ20170412170118573).

%
%
%
\bibliographystyle{splncs04}
\bibliography{paper783}

\end{document}